\documentclass{article}
\usepackage{jmlr2e}
\firstpageno{1}%

\usepackage{lastpage}

\usepackage[utf8]{inputenc}
\inputencoding{utf8}

\usepackage{booktabs}
\usepackage{tikz}
\usepackage{multirow}
\usepackage{graphicx}
\usepackage{cleveref}
\usepackage{listings}
\usepackage{wrapfig}

\newcommand{\pyes}{p_{yes}}
\newcommand{\pno}{p_{no}}
\newcommand{\indep}{\perp \!\!\! \perp}
\newcommand{\dep}{\not\perp\!\!\! \perp}

\begin{document}

\title{Large Language Models for Constrained-Based Causal Discovery}

 \author{\name{Kai-Hendrik Cohrs}\email{kai.cohrs@uv.es}\\
 \name{Gherardo Varando} \email{gherardo.varando@uv.es}\\
  \name{Emiliano Diaz} \email{emiliano.diaz@uv.es}\\
   \name{Vasileios Sitokonstantinou} \email{vasileios.Sitokonstantinou@uv.es}\\
      \name{Gustau Camps-Valls} \email{gustau.camps@uv.es}\\
 \addr Image Processing Laboratory, Universitat de València, Spain
 }
\editor{}

\maketitle

\begin{abstract}
Causality is essential for understanding complex systems, such as the economy, the brain, and the climate. Constructing causal graphs often relies on either data-driven or expert-driven approaches, both fraught with challenges. The former methods, like the celebrated PC algorithm, face issues with data requirements and assumptions of causal sufficiency, while the latter demand substantial time and domain knowledge. This work explores the capabilities of Large Language Models (LLMs) as an alternative to domain experts for causal graph generation. We frame conditional independence queries as prompts to LLMs and employ the PC algorithm with the answers. The performance of the LLM-based conditional independence oracle on systems with known causal graphs shows a high degree of variability. We improve the performance through a proposed statistical-inspired voting schema that allows some control over false-positive and false-negative rates. Inspecting the chain-of-thought argumentation, we find causal reasoning to justify its answer to a probabilistic query. We show evidence that knowledge-based CIT could eventually become a complementary tool for data-driven causal discovery. 
\end{abstract}


\section{Introduction}

Understanding causality is imperative across various disciplines, as it offers critical insights into the mechanisms of complex systems. For example, in the field of epidemiology, recognizing the causal link between smoking and lung cancer has led to effective public health campaigns and anti-smoking policies, significantly reducing the prevalence of smoking-related diseases \citep{cornfield1959smoking,wynder1997tobacco}. 

Investigating causality enhances our understanding of the underlying processes and enables us to make informed decisions and take proactive measures to address pressing challenges.
Investigating causality poses considerable challenges, with constructing causal graphs representing a formidable task. Data-driven causal discovery methods, including prominent techniques like PC~\citep{PCalgo} and GES~\citep{Chickering2002}, encounter a range of issues. These methodologies rely heavily on large amounts of data, necessitating complex conditional independence tests that can be particularly challenging, especially when working with diverse and mixed data types. The assumption of causal sufficiency, which presumes that all relevant variables are observed, can lead to erroneous conclusions, especially when unobserved variables act as potential confounders between system variables. Notably, there exist alternative methods such as LPCMCI \citep{gerhardus2020high}, FCI \citep{PCalgo}, SVAR-FCI \citep{malinsky2018causal}, and GPS \citep{claassen22a} that do not assume causal sufficiency~\citep{camps2023discovering}. Nonetheless, missing data and selection bias continue to pose persistent challenges in real-world applications, prompting efforts to develop more resilient causal discovery methods~\citep{camps2023discovering}.

In addition to data-driven causal discovery methods, another approach for creating causal graphs involves leveraging domain knowledge. However, this process is inherently challenging and time-consuming, demanding substantial expertise and labor~\citep{long2023can}. Experts tasked with constructing causal graphs must possess a deep understanding of the relationships and mechanisms within the system under investigation, which often requires knowledge across different disciplines. This often entails extensive consultations, discussions, and reviews with domain specialists, adding significant time and resource commitments. Furthermore, the complexity of many real-world systems amplifies the difficulty of accurately capturing all relevant causal relationships, leading to potential oversights and inaccuracies in the resulting causal graph. These challenges underscore the necessity for more automated methodologies. In that respect, LLMs could play a key role if they prove to be a reliable source of causal knowledge.

LLMs present a promising knowledge-driven alternative to expert-based graph building or data-driven causal discovery methods. \if Their unique emergent qualities enable them to distil existing knowledge. \fi They have shown good performance across a range of language understanding and logical reasoning tasks \citep{fewShotLearners,xu2023large}. This could extend to probabilistic and causal reasoning, including interventional and counterfactual scenarios. Whether or not this is the case is the subject of heated debate \citep{hobbhahn2022investigating, willig2023causal, zevcevic2023causal}. Despite the inherent complexity of directly asking LLMs to provide complete causal graphs, given the various levels of reasoning involved and the rich context necessary, different strategies exist to elicit causal graphs from LLMs effectively. These strategies include LLMs to predict causal relations~\citep{kiciman2023causal,long2023causal, willig2023causal}, LLMs as priors for data-driven causal discovery methods~\citep{ban2023query}, and LLMs to aid in downstream causal inference tasks by predicting the causal order of variables~\citep{vashishtha2023causal}.

To marry the traditional and the LLM-based methods, our work proposes \textit{chatPC} as a hybrid approach. Specifically, this work:
\begin{itemize}
    
    \item Frames conditional independence queries as prompts to LLMs and employs the PC algorithm with this oracle for causal graph construction.
    
    \item Evaluates the performance of LLMs on conditional independence tests across various problems, showing varied performance.
    
    \item Introduces a statistical-based approach for aggregating multiple LLM answers, improving performance.
    
    \item Examines the graphs predicted by PC with the LLM oracle, finding them to be reasonable.
    
    \item Identifies a general tendency for conservative answers from LLMs compared to experts and finds traces of causal reasoning in the model's answers. 
    
    \item Suggests that the approach could mitigate limitations of traditional methods, offering a promising avenue for automated causal graph construction.
\end{itemize}
We argue that relying solely on LLM-based bivariate causal queries inherently overlooks the presence of mediators, consequently limiting the comprehensive understanding of the full causal graph. Additionally, our research explores the extent to which LLMs' queries can effectively substitute data-driven conditional independence tests, considering that PC represents the current state of the art in causal discovery, and under a perfect oracle, it can consistently retrieve the ideal graph up to the Markov equivalence class \citep{PCalgo}. 

\begin{figure}
\centering
     \begin{tikzpicture}
      \node (Z) at (0,0) {$Z$};
      \node (Y) at (2,0) {$Y$};
      \node (W) at (2,2) {$W$};
      \node (X) at (0,2) {$X$};
      \node[draw=none] (O1) at (2.5,1.5) {};
      \node[draw=none] (O2) at (2.5,0.5) {};
      \node[draw=none] (O3) at (4.5,1.5) {};
      \node[draw=none] (O4) at (4.5,0.5) {};

    \node[below right] at (4.6,2.3) {\includegraphics[width=2cm, height=2cm]{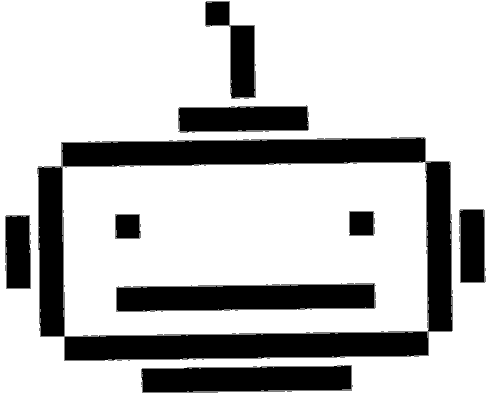}};

      \draw (X) -- (Z);
      \draw (Y) -- (Z);
      \draw (Y) -- (W);
      \draw (W) -- (Z);
      \draw[color=red] (X) -- (Y);

      \draw[->, bend left] (O1) to node[pos=0.5, above]{$X\indep Y|Z$?} (O3);
      
      \draw[->, bend left] (O4) to node[pos=0.5, below]{YES} (O2);
    \end{tikzpicture}
  \caption{Illustration of the introduced scheme for PC with GPT/LLM. {\sf \small Credits: Little robot face by Antònia Font.}}
\label{fig:scheme}
\end{figure}
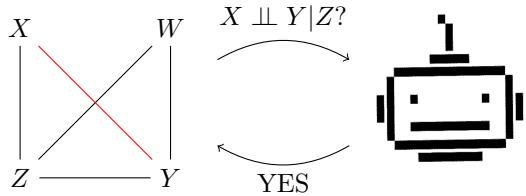

\section{Background}\label{sec:background}

This section presents the causal discovery task and briefly reviews different methodologies. We start by describing classical data-driven causal discovery methods, briefly review the integration of data and prior knowledge, and present novel approaches to causal discovery using LLMs.

\subsection{Data-Driven Causal Discovery}
Nowadays, there exists solid and established literature on causal discovery, which is the task of generating and selecting appropriate causal hypotheses from data. 
Starting from the seminal work of \citet{wright1934method}, using graphs to represent causal relationships has become widespread in causal analysis.
In the prevalent structural equation model (SEM) formalism, a causal model is described by a system of equations that implicitly define a directed acyclic graph (DAG) among the variables of the system through the signature of each equation~\citep{peters2017elements}. 
Importantly, in this framework, a causal model entails a statistical or data-generating process of the data while representing additional information such as interventional and counterfactual quantities~\citep{peters2017elements}. 
Causal discovery methods in the statistical DAG framework are usually categorized as constrained-based or score-based approaches~\citep{glymour2019review, SCUTARI2019235}.
The former methods attempt the search for graphs that most closely entail the set of conditional independence relations judged to hold in the population, while the latter approaches search the spaces of graphs for the best scoring ones for some score function (e.g., penalized likelihood, such as BIC or AIC scores).
A famous example of score-based methods is the GES algorithm~\citep{meek1997graphical, Chickering2002, chickering2002learning}, which performs simple moves (add/remove edges, invert directions) directly in the space of Markov equivalence classes. 
The most well-known constrained-based method is the PC-algorithm~\citep{PCalgo}, which we briefly describe as it will be central in our successive examples of LLM's integration. 
It follows three basic steps: (i) in the first step conditional independence statements ($X \indep Y | Z$), obtained through hypothesis tests performed on the data, are used to prune the edges of the fully connected undirected graph, thus building the so-called skeleton of the causal graph. (ii) In the second step, colliders are oriented such that invalid conditional (in)dependencies are ruled out. (iii) The third and last step applies a set of graphical rules~\citep{meek1995} to orient as many edges as possible. 
Both the GES and the celebrated Peter-Clark (PC) algorithm~\citep{PCalgo} have been proven consistent in the large sample limit under specific assumptions. In particular, if we consider an \emph{oracle} for conditional independence statements, and we assume causal sufficiency and faithfulness, the PC algorithm outputs a partially directed graph that represents the true Markov equivalence class (without additional assumption, it is impossible, from statistical information alone, to recover something more then the MEC).
We refer the readers to \citet{pearl2000models},  \citet{peters2017elements} and recent surveys and reviews~\citet{camps2023discovering, glymour2019review, SCUTARI2019235} for a more exhaustive overview of causal discovery algorithms and the DAG/SEM formalism.

\subsection{Causal Discovery with Prior Knowledge}
\label{sec:dataexpert}

The need for integrating prior expert knowledge and data-driven methods has been recognized since initial works on causal discovery~\citep{constantinou2023impact}. 
The Bayesian methodology straightforwardly incorporates prior knowledge with data-driven approaches~\citep {heckerman1995learning,friedman2003being, xu2015structure, bidag}. 
Methods have been developed to build informative priors on graphs based on agreement with prior beliefs of different types~\citep{mukherjee2008network} or with partial prior knowledge~\citep{castelo2000priors}.
Moreover, the Bayesian paradigm has also been used to integrate data with multiple sources of prior knowledge~\citep{amirkhani2016exploiting,werhli2007reconstructing}.
Prior knowledge integration has also been proposed and employed in non-Bayesian settings. 
Classical score-based and constrained-based algorithms can be combined with different prior knowledge such as structural restrictions~\citep{de2007bayesian, de2011efficient,li2018bayesian}
or information on directed causal paths~\citep{borboudakis2012incorporating,borboudakis2013scoring}.
Similarly, prior knowledge given as ancestral or topological order constraints can help guide exact methods~
\citep{chen2016learning,wang2021learning}.
Alternatively, human-in-the-loop approaches perform active interactions with the experts to guide the data-driven methods in discovering the causal graph~\citep{masegosa2013interactive, dasilva2023humanintheloop}.

\subsection{Causality with LLMs}
\label{subsec:causalLLM}
LLMs present a promising knowledge-driven alternative to expert-based graph building or data-driven causal discovery methods. Different strategies exist to elicit causal graphs from LLMs effectively. These strategies include LLMs to predict causal relations~\citep{kiciman2023causal,long2023causal, willig2023causal}, LLMs as priors for data-driven causal discovery methods~\citep{ban2023query}, and LLMs to aid in downstream causal inference tasks by predicting the causal order of variables \citep{vashishtha2023causal}.  
\citet{kiciman2023causal} showcase LLMs' knowledge-based causal discovery capabilities by eliciting DAGs for a series of bivariate and multivariate benchmark datasets where the ground truth is known. \citet{vashishtha2023causal} proposed an approach using the topological order over graph variables, showcasing the potential of LLMs in obtaining causal order and enhancing causal inference in diverse fields. \citet{long2023can} and \citet{long2023causal} examined the role of LLMs in facilitating the construction of causal graphs, focusing on the automation of edge scoring and the use of expert knowledge to improve data-driven causal graph identification, respectively.
\citet{zevcevic2023causal} argued against the causal capabilities of LLMs, defining a subgroup of Structural Causal Models (SCMs) and emphasizing the reliance of LLMs on embedded causal knowledge. \citet{cox2023causal} conducted a Socratic dialogue with an LLM, highlighting the necessity of sustained questioning for refining human reasoning imitated by LLMs and the reliability of their conclusions. \citet{pawlowski2023answering} and \citet{gat2023faithful} explored the use of LLMs in enabling precise causal reasoning and model-agnostic explanations. \citet{chen2023mitigating} focused on Bayesian Networks (BNs) and proposed strategies for addressing erroneous prior causal statements in LLM-based causal structure learning.  \citet{zhang2023causality} discussed the potential of enhancing causal machine learning and leveraging LLMs for improved performance and trustworthiness.

\section{Conditional Independence Queries via LLM}

We start by describing and evaluating conditional independence (CI) queries with LLMs. Specifically, we are interested in estimating the validity of conditional independence statements of the type:
\begin{quote}
    \emph{Is $X$ independent of $Y$ given $Z=(Z_1, Z_2, \ldots, Z_k)$?},
\end{quote}
without having access to observations of the involved variables. 
Instead, we would like to rely 
on \textit{available or expert} knowledge 
accessible through LLMs. 

For an LLM to be able to answer CI queries, it needs to be 
presented with some context and additional information related to the variables of interest alongside their description. 
In particular, we assume that for each \textit{problem}, we have access to the following information:
\begin{description}
    \item[\textbf{variables}] names or acronyms plus a short description for each quantity of interest.
    \item[\textbf{field}] the general subject area or expert field related to the problem.
    \item[\textbf{context}] a description of the broad context of the variables under consideration, including relevant details that go beyond general knowledge
\end{description}

\subsection{Prompting for conditional independence testing}

While the approach we propose in this work could be implemented with any LLM trained to follow instructions, we compare GPT-3.5-turbo\footnote{\url{https://platform.openai.com/docs/models/gpt-3-5}} and GPT-4\footnote{\url{https://platform.openai.com/docs/models/gpt-4-and-gpt-4-turbo}} from OpenAI. The former is a less computationally-expensive model with good performance over different benchmarks, while the latter is among the current state-of-the-art LLMs~\citep{chiang2024chatbot}.
For a CI statement ($X\indep Y|Z$), we consider a simple prompt that
combines the field and context information and a description of the involved variables ($X, Y$, and eventually $Z$) with a general instruction and 
response template as follows (see \cref{sec:prompt} for a detailed specification of the used prompt):
\begin{verbatim}
    Persona specification
    Instructions
    Context
    Variables description
    CI Statement question
    Response template
\end{verbatim}
The persona is based on the field variable and primes the LLM to produce reasoning appropriate to the area under study. To improve the answers, we apply chain-of-thought prompting following \citet{wei2023chainofthought}, enabling us to gain insights into the model's reasoning and inspect if it is causally inspired. Further, we ask the model to provide uncertainty about its best guess along the lines of \citet{tian2023just}. 

\subsection{Testing}
Various strategies can be followed in order to perform a ``Hypothesis test'' for a conditional independence statement with LLMs. A naive option consists of asking the LLM a single question with the 
prompt described in the previous section and decide that 
a statement is valid (i.e., the variables are indeed independent) if the answer is YES and invalid, if it is NO.
The main problem with this approach is that since LLMs are probabilistic, 
a single answer from an LLM does not need to correspond to its mode (the most likely answer) or could fail to respect the required response template: its answer could be, for instance, UNCERTAIN instead of YES/NO. 
Instead, we ask for an independent batch (size $n$) of answers, parse the obtained answers (YES or NO) together with the reported uncertainties, and finally output either an answer based on simple voting or weighted voting where the weights are the reported probabilities.

Alternatively, we implement a ``statistical approach'' where we actually produce $p$-values for the null hypothesis $\pno\geq\pyes$ (or alternatively $\pno \leq \pyes$) where $\pno, \pyes$ are the proportion of NO and YES answers over the total requested batch $n$. 
The constructed test is based on the idea that we want to test whether the probability of obtaining the answer NO is significantly higher than that of obtaining the answer YES and vice versa. We opt for the null hypothesis if this difference is not significant in light of the obtained responses.
A final decision is obtained by setting a 
significance level $\alpha$ and rejecting the chosen null hypothesis if the p-value is less or equal to $\alpha$ (we will employ $\alpha=0.05$ in the experiments).
This last strategy has the advantage of considering the random variability of the answers and 
could offer a principled way of controlling the false positive rate. 
The user could then specify, for a particular problem, which of the two null hypotheses they would like to employ (either $\pno \geq \pyes$ or $\pyes \geq \pno$), which in turn implies a different false-positive control (considering either NO or YES as positive).

\subsection{Evaluation}

We evaluate the performance for CI testing on various problems defined in the BNLearn repository~\citep{Scutari2014}, the spurious correlation website~\citep{vigen}, a classical problem on reconstructing protein-signaling networks~\citep{sachs2005} and a set of graphs on climatic teleconnections~\citep{Kretschmer2021}: 
\begin{itemize}
    \item \texttt{cancer} Simple causal graph involving four factors influencing the probability of cancer~\citep{korb2010bayesian}.
    \item \texttt{burglary} The classical earthquake example in~\citep{korb2010bayesian, Scutari2014} 
    \item \texttt{asia} Causal graph of eight factors linked to respiratory problems~\citep{lauritzen1988local}. 
    \item \texttt{sachs}: Causal graph among $11$  phosphorylated proteins  and phospholipids in single-cell data~\citep{sachs2005}.
    \item \texttt{spurious}: Famous examples of spuriously correlated pairs of variables obtained from the spurious-correlation website~\citep{vigen}.
    \item \texttt{bk-spv}: Hypothesized pathway from autumn sea ice in the Barents and Kara Seas (BK) to the stratospheric polar vortex (SPV) in winter, including various confounding and mediating factors~\citep{Kretschmer2021}.
    \item \texttt{nao-dk-med}: Hypothesized causal relation between summertime North Atlantic Oscillation (NAO) and precipitation patterns in Denmark (DK) and the Mediterranean (MED)~\citep{Kretschmer2021}.
\end{itemize}
For all graphs except \texttt{sachs}, we evaluate all the possible CI statements over all variables with both permutations of $X$ and $Y$. 
For \texttt{sachs}, we evaluate all CI statements (with $X$-$Y$ permutations) up to conditioning sets of size $1$.
Lastly, for \texttt{spurious}, we evaluate the 
marginal independence statements $X \indep Y$ for all pairs of variables. 
For all the experiments, we obtain the answer to the CI queries by aggregating, as described previously, $n=10$ independent batched responses from the LLM.
The code implementing our approach and the problem descriptions are available in the repository \url{https://github.com/IPL-UV/causal_GPT}.

\begin{table}
\centering
\caption{Performance of LLM-based conditional independence tests with different voting procedures and metrics (accuracy, F1, precision, and recall.) The cells are empty when the metric cannot be computed.}
\label{tbl:performance}
\setlength{\tabcolsep}{3pt}

    \begin{tabular}{llccccc}
        \toprule
        \textbf{Data} & \textbf{Model} & \textbf{Method} & \textbf{Accuracy} & \textbf{F1} & \textbf{Precision} & \textbf{Recall} \\ 
        \midrule
        \multirow{6}{*}{\texttt{cancer}} & \multirow{3}{*}{GPT-3.5} & voting & 0.64 & 0.39 & 0.36 & 0.43 \\ 
        &  & stat. Test ($H_0$: $\dep$)  & 0.72 & 0.08 & 0.05 & 0.33 \\ 
        &  & stat. Test ($H_0$: $\indep$) & 0.47 & \textbf{0.42} & \textbf{0.71} & 0.29 \\ 
        \cmidrule(lr){2-7}
        & \multirow{3}{*}{GPT-4} & voting & 0.72 & 0.25 & 0.17 & \textbf{0.54} \\ 
        &  & stat. Test ($H_0$: $\dep$)  & \textbf{0.73} & 0.19 & 0.12 & 0.46 \\ 
        &  & stat. Test ($H_0$: $\indep$) & 0.70 & 0.31 & 0.26 & 0.39 \\ 
        \midrule
        \multirow{6}{*}{\texttt{burglary}} & \multirow{3}{*}{GPT-3.5} & voting & 0.45 & 0.36 & 0.50 & 0.28 \\ 
        &  & stat. Test ($H_0$: $\dep$)  & 0.62 & 0.27 & 0.26 & 0.28 \\ 
        &  & stat. Test ($H_0$: $\indep$) & 0.38 & 0.43 & 0.90 & 0.28 \\ 
        \cmidrule(lr){2-7}
        & \multirow{3}{*}{GPT-4} & voting & 0.81 & 0.77 & \textbf{1.00} & 0.63 \\ 
        &  & stat. Test ($H_0$: $\dep$)  & \textbf{0.88} & \textbf{0.82} & \textbf{1.00} & \textbf{0.69} \\ 
        &  & stat. Test ($H_0$: $\indep$) & 0.79 & 0.72 & \textbf{1.00} & 0.56 \\ 
        \midrule
        \multirow{6}{*}{\texttt{asia}} &  &
        data-driven (MI)  & 0.58 & {0.56} & 0.54 & 0.58 \\ 
         \cmidrule(lr){2-7}
        & \multirow{3}{*}{GPT-3.5} & voting & 0.51 & 0.39 & 0.36 & 0.43 \\ 
        &  & stat. Test ($H_0$: $\dep$)  & {0.61} & 0.18 & 0.12 & 0.44 \\ 
        &  & stat. Test ($H_0$: $\indep$) & 0.48 & 0.52 & \textbf{0.75} & 0.39 \\
        \cmidrule(lr){2-7}
        & \multirow{3}{*}{GPT-4} & voting & 0.73  & 0.63  & 0.56 & 0.71  \\ 
        &  & stat. Test ($H_0$: $\dep$)  & 0.74  & 0.58  & 0.48 & \textbf{0.73} \\ 
        &  & stat. Test ($H_0$: $\indep$) & \textbf{0.76}  & \textbf{0.69}  & 0.72 & 0.66  \\ 
        \midrule
        \multirow{6}{*}{\texttt{sachs}} &  & data-driven~\citep{sachs2005} & 0.61 & \textbf{0.21} & 0.85 & {0.12} \\
        \cmidrule(lr){2-7}
        & \multirow{3}{*}{GPT-3.5} & voting & 0.45 & 0.13 & 0.53 & 0.08 \\ 
        &  & stat. Test ($H_0$: $\dep$)  & 0.81 & 0.13 & 0.24 & 0.09 \\ 
        &  & stat. Test ($H_0$: $\indep$) & 0.20 & 0.13 & \textbf{0.94} & 0.07 \\ 
        \cmidrule(lr){2-7}
        & \multirow{3}{*}{GPT-4} & voting & 0.93 &  &  &  \\ 
        &  & stat. Test ($H_0$: $\dep$)  & \textbf{0.94} &  &  &  \\ 
        &  & stat. Test ($H_0$: $\indep$) & 0.93 & 0.15 & 0.10 & \textbf{0.27} \\
        
        \midrule
        \multirow{6}{*}{\texttt{bk-spv}} & \multirow{3}{*}{GPT-3.5} & voting & 0.55 & 0.41 & 0.51 & 0.34 \\ 
        &  & stat. Test ($H_0$: $\dep$)  & 0.72 & 0.24 & 0.17 & 0.42 \\ 
        &  & stat. Test ($H_0$: $\indep$) & 0.41 & \textbf{0.44} & \textbf{0.90} & 0.27 \\ \cmidrule(lr){2-7}
        & \multirow{3}{*}{GPT-4} & voting & \textbf{0.75} &  &  \\ 
        &  & stat. Test ($H_0$: $\dep$)  & \textbf{0.75} &  &  \\ 
        &  & stat. Test ($H_0$: $\indep$) & 0.74 &  &  \\ 
        \bottomrule
    \end{tabular}
\end{table}

\begin{wrapfigure}{r}{0.6\textwidth}
\begin{tabular}{cc}
\label{fig}
   (a) GPT-3.5  & (b) GPT-4 \\
    \includegraphics[width=4cm]{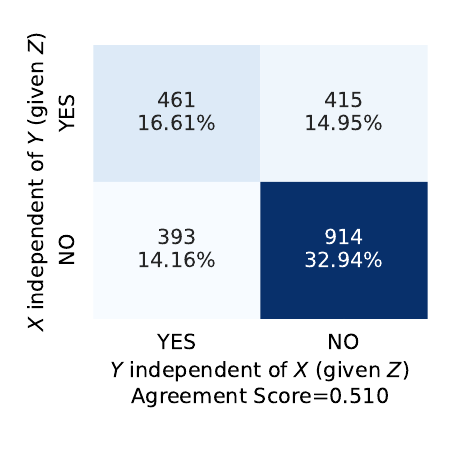}
& 
    \includegraphics[width=4cm]{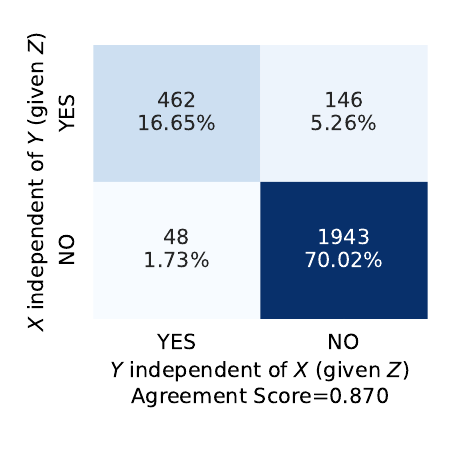} 
        \label{fig: consistency gpt-4}
\end{tabular}
   \caption{Confusion matrix of the model's responses to queries with changing order of $X$ and $Y$. UNCERTAIN outcomes in case of a tie in majority voting are hidden. The agreement score aggregates common YES, NO and UNCERTAIN outcomes.}
 \label{fig:consistency}
\end{wrapfigure}

\paragraph{Permutation consistency}
Consistency is a fundamental property that human experts are tested for and is shown to correlate with confidence and disagreement~\citep{litvinova2022experts}.
In our context, consistency means that the result of a conditional independence test should not depend on the order of variables, i.e., it should be commutative in $X$ and $Y$ (given $Z$). As a first sanity check, we checked the consistency of the responses across all CITs, with respect to the change of order of $X$ and $Y$ (see Figure \ref{fig:consistency}) for both language models.

For GPT-3.5, the agreement over the (conditional) independence tests reaches merely 51\%, with around 22\% of all answers being UNCERTAIN in at least one of the directions. This indicates that the results for GPT-3.5 vary substantially over multiple runs, particularly under a change of order of the variables. In contrast, for GPT-4, in 87\% of all cases, the answers for both directions of the CITs yielded the same result. Albeit with small deviations due to stochasticity, this is a satisfactory result in consistency for GPT-4. In contrast to GPT-3.5, however, the overall proportion of CIT answered NO was over 70\%. Although this proportion may be inflated by relatively large and dense graphs such as \texttt{sachs}, it indicates a tendency of GPT-4 towards rejecting independence as the safe answer.  
Given the remaining variation, especially in the responses of GPT-3.5, we propose to aggregate the query results in both directions to obtain results invariant to the order of the two involved variables.

\paragraph{Performance of CIT} 
Table~\ref{tbl:performance} summarizes the evaluations of the CI queries over the different problems. We compute the standard classification metrics (accuracy, precision, recall, and F1 scores). The results show varying performance over the different causal graphs and models. As some of these problems constitute pedagogical examples or hypothesized graphs, they are not accompanied by observations. Hence, we only compare against data-driven baselines for \texttt{sachs} and \texttt{asia}. 
Overall, we can see that the statistical tests, which imply certain assumptions about the graph's sparsity and require stronger evidence (i.e., more votes) for rejecting either dependence or independence, give some control over the metrics, usually implying a trade-off between recall and precision. 
For \texttt{cancer}, GPT-3.5 yields the best results in terms of F1 and precision, while GPT-4 has the higher accuracy. Inspection of the implied graph under the chatPC (see~\cref{fig:cancer}) suggests that GPT-4 assumes more dependencies and finds a denser graph.
The context of \texttt{burglary} contains mostly all information on how the variables are connected. The language models still need to perform the task of extracting valid CI statements from there. GPT-4 clearly outperforms GPT-3.5 in this regard. This suggests that given access to the relevant information, GPT-4 mostly makes correct statements about CI.
On the \texttt{asia} graph, the LLM models beat the data-driven baseline based on mutual information in all metrics. Only in precision, GPT-3.5 achieves a slightly higher value than GPT-4. 
On the \texttt{sachs} data, in terms of F1, the data-driven method~\citep{sachs2005} outperforms the two language models, whilst GPT-4 shows the highest accuracy. It is noteworthy that GPT-4 avoids eliminating the connection of the proteins Plc-$\gamma$ and PIP2, with PKC, two known mistakes of the data-driven approach. 
The LLM-driven CITs do not yield a perfect oracle in their current state. Some improvement can likely be made with the next generations of language models, fine-tuning of the prompt~\citep{cheng2023blackbox}, or disambiguation of the context. %
A final source of error is that the required knowledge may be controversial or may simply not exist.
Here, systems based on retrieval-augmented-generation (RAG)~\citep{lewis2021retrievalaugmented} could help to ground answers on secured knowledge or weigh sources against each other (old vs. new; unsystematic vs. scientific). 
As data-driven methods also reveal major failure modes, using knowledge-driven evidence to complement data can be a promising future research avenue.

\paragraph{Inquiring spurious correlations}
Up to this point, GPT-4 seemed overly conservative in favor of dependence.
To probe this further, we went to the other extreme and asked for statistical independence between variables taken from the spurious correlation website (see Appendix,  Table~\ref{tbl:spurious}). In contrast, in this case, the model almost always chose independence. Nevertheless, there are notable exceptions, such as per \textit{capita consumption of chicken} and \textit{US crude oil imports}, where the LLM correctly suggests that oil imports and, in turn, oil prices may affect chicken prices and its consumption through transportation costs. Furthermore, broader economic factors may serve as a common confounder (for the full response see \cref{sec: response 1}). Other slightly less clear connections seem to exist in the pairs \textit{total revenue generated by arcades - computer science doctorates}, and \textit{Japanese passenger cars sold in the US - suicides by crashing of motor vehicle}, where the voting almost resulted in stalemates. We may favour dependence or independence depending on our choice of the null hypothesis. 
It is important to note that data-driven methods would always detect dependence as the spurious correlation examples are selected in this manner. This can happen in any data by chance; Hence, consulting additional knowledge-based evidence as a safety net seems reasonable.

\section{Application to Causal Discovery}

We propose to couple the CI oracle or testing, introduced in the previous section, with the PC-algorithm~\citep{colombo14a} for recovery of the Markov equivalence class of a causal graph. 
The PC algorithm starts from a fully connected skeleton (undirected graph) among the considered variables and iteratively removes edges between variables $X$ and $Y$
when it finds a conditioning set $Z$ such that $X$ is independent of $Y$ given $Z$. After the so-called skeleton phase, v-structures are identified through specific CI testing, and finally, a set of orientation rules are applied~\citep{meek1995}.
Hence, to implement chatPC, we plug in 
the LLM-based conditional independence testing to an existing PC implementation~\citep{ATIENZA2022}.

\paragraph{Causal graphs from the examples}
The PC algorithm should not be run only with LLMs for arbitrarily large graphs. First, CITs with growing conditioning set size turn even for an expert increasingly into mental gymnastics as you technically verify d-separation. Second, with the increasing number of variables, the number of required CITs grows exponentially. This renders the approach under the current energy and financial costs for querying LLMs infeasible unless the graph is small or one couples it with data-driven CITs.
We applied chatPC to the \texttt{burglary}, \texttt{cancer}, \texttt{asia} and \texttt{nao-dk-med}.

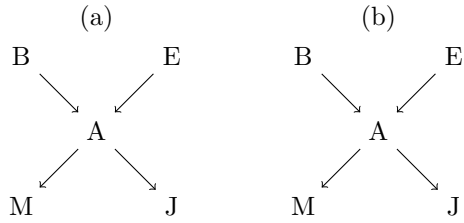
\begin{figure}
    \centering
    \begin{tikzpicture}
    \node
        (aa) at (1,2.5) {(a)};
        \node (B)  at (0,2) {B};
        \node (E)  at (2,2) {E};
        \node (A)  at (1,1) {A};
        \node (J)  at (2,0) {J};
        \node (M)  at (0,0) {M};
        \draw[->] (E) -- (A);
        \draw[->] (B) -- (A);
        \draw[->] (A) -- (J);
        \draw[->] (A) -- (M);
    \end{tikzpicture}
    \hspace{1cm}
    \begin{tikzpicture}
        \node (b) at (1,2.5) {(b)};    
        \node (B)  at (0,2) {B};
        \node (E)  at (2,2) {E};
        \node (A)  at (1,1) {A};
        \node (J)  at (2,0) {J};
        \node (M)  at (0,0) {M};
        \draw[->] (E) -- (A);
        \draw[->] (B) -- (A);
        \draw[->] (A) -- (J);
        \draw[->] (A) -- (M);
    \end{tikzpicture}
    \caption{Assumed true graph (a) and skeleton recovered (b) with the proposed chatPC approach for the \texttt{burglary} problem.
    Variables: Burglary in progress (B); earthquake (E); alarm ringing (A); Mary (M) or John (J) calling.}
    \label{fig:burglary}
\end{figure}

\begin{figure}
    \centering
    \begin{tikzpicture}
        \node
        (aa) at (1,2.5) {(a)};
        \node (P)  at (0,2) {P};
        \node (S)  at (2,2) {S};
        \node (C)  at (1,1) {C};
        \node (D)  at (2,0) {D};
        \node (X)  at (0,0) {X};
        \draw[->] (P) -- (C);
        \draw[->] (S) -- (C);
        \draw[->] (C) -- (X);
        \draw[->] (C) -- (D);
    \end{tikzpicture}
    \hspace{1cm}
    \begin{tikzpicture}
        \node (b) at (1,2.5) {(b)};
        \node (P)  at (0,2) {P};
        \node (S)  at (2,2) {S};
        \node (C)  at (1,1) {C};
        \node (D)  at (2,0) {D};
        \node (X)  at (0,0) {X};
        \draw[-] (P) to[bend right] (D);
        \draw[-] (S) -- (D);
        \draw[-] (P) -- (C);
        \draw[-] (S) -- (C);
        \draw[-] (C) -- (X);
        \draw[-] (C) -- (D);
    \end{tikzpicture}
    \caption{Assumed true graph (a) and skeleton recovered (b) with the proposed chatPC approach for the \texttt{cancer} problem.
    Variables: Patient is a smoker (S); patient exposed to pollution (P); patient suffers from lung cancer (C); positive results from a chest X-ray (X); patient is suffering from dyspnoea (D).}
    \label{fig:cancer}
\end{figure}
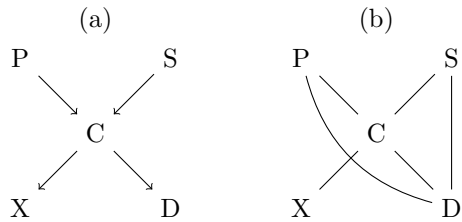

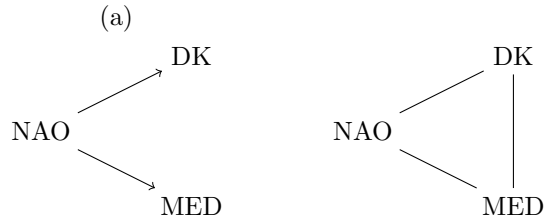
\begin{figure}
    \centering
    \begin{tikzpicture}
        \node
        (aa) at (1,2.5) {(a)};
        \node (NAO)  at (0,1) {NAO};
        \node (DK)  at (2,2) {DK};
        \node (MED)  at (2,0) {MED};
        \draw[->] (NAO) -- (DK);
        \draw[->] (NAO) -- (MED);
    \end{tikzpicture}
    \hspace{1cm}
    \begin{tikzpicture}
        \node (NAO)  at (0,1) {NAO};
        \node (DK)  at (2,2) {DK};
        \node (MED)  at (2,0) {MED};
        \draw[-] (NAO) -- (DK);
        \draw[-] (NAO) -- (MED);
        \draw[-] (DK) -- (MED);
    \end{tikzpicture}
    \caption{Assumed true graph (a) and skeleton recovered (b) with the proposed chatPC approach for the \texttt{nao-dk-med} problem.
    Variables: North-Atlantic Oscillation (NAO); summer precipitation in Denmark (DK); summer precipitation in the Mediterranean region (MED).}
    \label{fig:nao-dk-med}
\end{figure}

First, for \texttt{burglary}, chatPC can retrieve the correct assumed causal graph (see \cref{fig:burglary}). This result stands even though the method made some mistakes in the CITs for \texttt{burglary}. The graph of the \texttt{cancer} variables (see \cref{fig:cancer}) contains the right sub-graph, which means that the LLM introduced no incorrect independence. The links between pollution and dyspnoea and smoking and dyspnoea are absent in the ground truth, but research suggests that such pathways might be assumed~\citep{Berliner2016}. An expert may choose to introduce sparsity into a graph by including only the strongest links. With LLM-based CITs, this is possible through the significance level of the statistical tests, which we could tune up to the point where some of these links disappear. It is also noteworthy that chatPC could not direct any edge in the retrieved graph. In principle, directing edges is a simpler task that could, in a second step, also be carried out with LLMs~\citep{kiciman2023causal}. For \texttt{asia} (see \cref{fig:asia}), we find an overall reasonable graph that almost matches the skeleton of the ground truth. Notable differences are that the asia node is disconnected, smoke is directly connected to dysp, which aligns with the \texttt{cancer} graph, and xray only has a direct link to bronc. 
Finally, in \texttt{NAO-DK-MED}, the hypothesized graph suggests that the summertime North-Atlantic Oscillation is the single common cause of precipitation patterns in Denmark and the Mediterranean. The LLM would hence need to identify that DK $\indep$ MED$|$ NAO, but it fails to do so (see \cref{fig:nao-dk-med}). \citet{Kretschmer2021} assume that the data is detrended, and even then, it is a debatable assumption that there are no connections left. Hence, the graph retrieved by chatPC is arguably not false. Moreover, one can see some signal reflecting this as, in contrast to all other CITs showing 20-0 (NO-YES proportions) for dependence, DK $\indep$ MED$|$ NAO resulted in 17-3 (NO-YES). If one were to require a complete agreement for dependency, the graph would have matched the assumed ground truth.

\section{Conclusions}
Our work contributes to the existing literature by probing an alternative to the data-driven PC approach, leveraging the capabilities of LLMs for PC when data is limited or unavailable. 
Building a reliable knowledge-based conditional independence \textit{oracle} could either provide a prior to constrain its data-driven counterpart or deliver a substitute for data-driven methods. Our analysis attempts to shed light on where we stand in this endeavour.

We found that LLM conjectures about hidden confounders, showing that they use causal reasoning to tackle this primarily statistical task. This, however, is neither consistently nor always successfully done. The varying performance over different tasks showed that more effort is needed to steer the models to more efficient causal reasoning. We proved that employing an aggregating mechanism framed as a statistical test leads to improved performance and effective control over the retrieved graph's density in the context of the PC algorithm.
The causal graphs predicted by the PC algorithm with LLM-based CITs appear reasonable. While not infallible, the method demonstrates potential in capturing meaningful causal relationships, offering a promising avenue for automated causal graph construction in combination with data-driven methods.
Finally, we found that LLMs generally tend toward conservative answers, contrasting with the often bolder responses from human experts. Understanding and addressing the cautious nature of LLM reasoning is crucial for refining the accuracy and reliability of the generated causal graphs.
Future work will explore the combination of data-driven and language-driven causality, where relying on CIT estimates in PC schemes constitutes a sound framework to seamlessly integrate evidence based on data and knowledge. Here, RAG-based systems~\citep{lewis2021retrievalaugmented} could play a key role by ensuring that retrieved knowledge stems from scientific sources. 
The proposed approach could be useful in \emph{alignment} and evaluating LLMs. Probing CITs and their implication for a causal graph construction with LLMs is useful to understand the level of causal reasoning and understanding of such systems. It does not escape our notice that, despite the many challenges in our pathway to causal reasoning with language models, the approach can bring many opportunities for scientific discovery.

\acks{This work was supported by the BBVA with the project \href{https://www.fbbva.es/noticias/concedidas-5-ayudas-a-equipos-de-investigacion-cientifica-en-big-data/}{`SCALE'}, Microsoft with the \href{https://www.microsoft.com/en-us/research/collaboration/microsoft-climate-research-initiative/projects/}{Causal4Africa} project and the GVA \href{https://isp.uv.es/ai4cs}{PROMETEO} AI4CS project (CIPROM/2021/056.)}

\bibliography{main}

\appendix
\section{Details on the Prompt}
\label{sec:prompt}
Here, we describe the details of the prompt used. 
In the following prompt scheme, curly brackets \verb|{}| denotes elements that are replaced 
by corresponding elements from the problem description and the specific CI statement which is queried.
In particular \verb|{field}| and \verb|{context}| are replaced by their values for the problem, and  \verb|{x}|, \verb|{y}| and \verb|{z}| are the name of the variables involved in the CI ($X \indep Y | Z$).
If $Z = \emptyset$ the whole 
\verb|conditioned on {z}| is dropped from the prompt.

{\footnotesize
\begin{verbatim}
    system: You are a helpful expert in {field} willing to answer
    questions. 
    You will be asked to provide your estimate and confidence
    on statistical independence between two variables 
    (eventually conditioned on
    a set of variables).
    Your answer should not be based on data or observations,
    but only on the available knowledge.
    Even when unsure or uncertain, provide a valid answer and uncertainty.
    Answer only in the required format. 

    user:
    {context}
    Consider the following variables:
    {variables list}
    is {x} conditionally independent of {y} conditioned on {z}?
    
    system: After explaining your reasoning, provide the answer between 
    brackets as YES/NO, with percentage uncertainty between parenthesis.
    Where YES stands for
    \"{x} is conditionally independent of {y} conditioned on {z}" 
    and NO stands for
    "{x} is not conditionally independent of {y} conditioned on {z}".
    For example [NO (50%
\end{verbatim}
}

\section{Spurious Correlations}
We report in Table~\ref{tbl:spurious} the 
responses obtained for the \texttt{spurious} 
problem over 15 marginal independence statements for the corresponding pairs of 
spuriously associated variables. 
We report the decisions obtained with the voting and the two statistical approaches; moreover, we report the number of NO and YES answers among the $n=20$ batched responses.
\begin{table}[h!]
\centering
\setlength{\tabcolsep}{1.5pt}
\caption{Predictions for variable pairs of the spurious correlations dataset with gpt-4. Numbers below each statistical test denote the respective $p$-value.}
\label{tbl:spurious}
\begin{tabular}{llccc}
\toprule
Variable & Description & voting & stat. Test & Test\\
 Name & & \small NO - YES & ($H_0: \indep$) & ($H_0: \not\indep$)\\
\midrule
spending & US spending on science, space, and technology & YES & YES & YES \\
suicides & Suicides by hanging, strangulation and suffocation & \small 4 - 16  & \small 1 & \small $\approx 10^{-4}$\\
\midrule
pool & number of people who drowned by falling into a & YES & YES & YES \\ & pool per year & \small 0 - 20  & \small 1  & \small $\approx 10^{-10}$ \\
cage & number of films Nicolas Cage appeared in per year &  \\
\midrule
cheese & per capita cheese consumption & YES & YES & YES\\
bed & number of people who died by becoming tangled in & \small 0 - 20 & \small 1 & \small $\approx 10^{-10}$\\ & their bedsheet & 
\\
\midrule
divorce & divorce rate in Maine & YES & YES& YES\\
margarine & per capita consumption of margarine & \small 1 - 19 &  \small 1 & \small $\approx 10^{-9}$ \\
\midrule
age & age of Miss America & YES & YES& YES\\
murder & number of people murdered by steam, hot vapors & \small 0 - 20 & \small 1 & \small $\approx 10^{-10}$ \\ & and hot objects &   \\
\midrule
revenue & total revenue generated by arcades & YES & YES & NO \\
CS & computer science doctorates awarded in the US & \small 9 - 11 & \small 0.74 & \small 0.26 \\
\midrule
launches & worldwide non-commercial space launches & YES & YES& YES \\
Soc & sociology doctorates awarded (US) & \small 3 - 17 & \small 1 & \small $\approx 10^{-6}$ \\
\midrule
mozzarella & per capita consumption of mozzarella cheese & YES& YES& YES \\
engineering & civil engineering doctorates awarded &  \small 0 - 19 & \small 1 & \small $\approx 10^{-10}$\\
\midrule
boat & people who drowned after falling out of a fishing boat & YES & YES& YES \\
Kentucky & marriage rate in Kentucky & \small 0 - 20 &\small 1 & \small $\approx 10^{-10}$  \\
\midrule
Norway & US crude oil imports from Norway & YES & YES& YES\\
railway & drivers killed in collision with railway train & \small 0 - 20 & 1 &\small $\approx 10^{-10}$ \\
\midrule
chicken & per capita consumption of chicken & NO & NO & NO \\
oil & US crude oil imports & \small 16 - 4 & \small $\approx10^{-4}$ &  \small 1 \\
\midrule
swimming& number people who drowned while in a & YES& YES& YES \\ -pool & swimming-pool & \small 0 - 20 & \small 1 &  \small $\approx 10^{-10}$  \\
power & power generated by US nuclear power plants & \\
\midrule
cars & Japanese passenger cars sold in the US & NO & YES& NO \\
crashing & Suicides by crashing of motor vehicle & \small 12 - 8 &  \small 0.1 & \small 0.9 \\
\midrule
spelling & letters in winning word of Scripps National & YES& YES& YES \\ & Spelling Bee & \small 0 - 20 &  \small 1  & \small $\approx 10^{-10}$\\
spiders & number of people killed by venomous spiders \\
\midrule
maths & math doctorates awarded & YES& YES& YES \\
uranium & uranium stored at US nuclear power plants & \small 0 - 20 & \small 1 & \small $\approx 10^{-10}$ \\
\bottomrule
\end{tabular}
\end{table}

\section{Response Example}
\label{sec: response 1}
{\footnotesize
\begin{verbatim}
    Considering the relationship between per capita consumption of chicken and
    US crude oil imports:
    
    - The per capita consumption of chicken might seem unrelated to crude oil 
    imports at first glance, as they belong to different sectors (agriculture 
    vs. energy).
    
    - However, there are indirect links between them. The price of oil can
    influence the cost of transportation and feed, which in turn can affect the 
    price of chicken and potentially consumption patterns.
    
    - Additionally, broader economic factors influenced by oil prices, such as 
    inflation rates and economic growth, could impact consumers'' purchasing
    power and preferences.
    
    Given these indirect connections, we cannot confidently state that there is
    absolute independence between these two variables. Therefore, the answer is
    [NO (75%
\end{verbatim}
}
\newpage
\section{Additional Results}
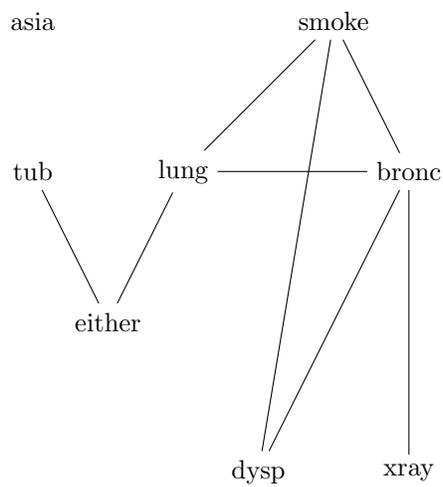
\begin{figure}
    \centering
    \begin{tikzpicture}[scale =2]
\node (asia) at (0,0) {asia};
\node (tub) at (0,-1) {tub};
\node (lung) at (1,-1) {lung};
\node (smoke) at (2,0) {smoke};
\node (bronc) at (2.5,-1) {bronc};
\node (dysp) at (1.5,-3) {dysp};
\node (xray) at (2.5,-3) {xray};
\node (either) at (0.5,-2) {either};
\draw[-] (bronc) -- (xray);
\draw[-] (smoke) -- (dysp);
\draw[-] (lung) -- (either);
\draw[-] (smoke) -- (bronc);
\draw[-] (bronc) -- (dysp);
\draw[-] (lung) -- (bronc);
\draw[-] (lung) -- (smoke);
\draw[-] (tub) -- (either);
\end{tikzpicture}
    \caption{Graph obtained with the chatPC approach on the \texttt{asia} problem. Variables: Patient has recently visited Asia (asia); patient has tuberculosis (tub); patient is a smoker (smoke); patient has lung cancer (lung); patient has bronchitis (bronc); patient has dyspnoea (dysp); patient has either tuberculosis or lung cancer or both (either); positive result in chest X-ray (xray).}
    \label{fig:asia}
\end{figure}
\end{document}